\documentclass[sigconf,dvipsnames]{acmart}

\copyrightyear{2023}
\acmYear{2023}
\setcopyright{acmlicensed}\acmConference[FIRE 2023]{Forum for Information Retrieval Evaluation}{December 15--18, 2023}{Panjim, India}
\acmBooktitle{Forum for Information Retrieval Evaluation (FIRE 2023), December 15--18, 2023, Panjim, India}
\acmPrice{15.00}
\acmDOI{10.1145/3632754.3632755}
\acmISBN{979-8-4007-1632-4/23/12}

\usepackage[T1]{fontenc}
\usepackage{graphicx}
\usepackage{hyperref}
\usepackage{xcolor}
\begin{document}

\title{Harnessing the Power of Prompt-based Techniques for Generating School-Level Questions using Large Language Models}

\author{Subhankar Maity}
\orcid{0009-0001-1358-9534}
\affiliation{%
 \institution{IIT Kharagpur}
      \state{West Bengal}
  \country{India}
}
\email{subhankar.ai@kgpian.iitkgp.ac.in}
\authornote{Corresponding Author}

\author{Aniket Deroy}
\orcid{0000-0001-7190-5040}
\affiliation{
 \institution{IIT Kharagpur}
\state{West Bengal}
\country{India}
}
\email{roydanik18@kgpian.iitkgp.ac.in}

\author{Sudeshna Sarkar}
\orcid{0000-0003-3439-4282}
\affiliation{%
 \institution{IIT Kharagpur}
 \state{West Bengal}
 \country{India}
 }
\email{sudeshna@cse.iitkgp.ac.in}

\begin{abstract}
Designing high-quality educational questions is a challenging and time-consuming task. In this work, we propose a novel approach that utilizes prompt-based techniques to generate descriptive and reasoning-based questions.
However, current question-answering (QA) datasets are inadequate for conducting our experiments on prompt-based question generation (QG) in an educational setting. Therefore, we curate a new QG dataset called \textit{\textit{EduProbe}} for school-level subjects, by leveraging the rich content of NCERT textbooks. We carefully annotate this dataset as quadruples of 1) \textit{Context}: a segment upon which the question is formed; 2) \textit{Long Prompt}: a long textual cue for the question (\textit{i.e.}, a longer sequence of words or phrases, covering the main theme of the \textit{context}); 3) \textit{Short Prompt}: a short textual cue for the question (\textit{i.e.}, a condensed representation of the key information or focus of the \textit{context}); 4) \textit{Question}: a deep question that aligns with the \textit{context} and is coherent with the \textit{prompts}. We investigate several prompt-based QG methods by fine-tuning pre-trained transformer-based large language models (LLMs), namely PEGASUS, T5, MBART, and BART. Moreover, we explore the performance of two general-purpose pre-trained LLMs such as Text-Davinci-003 and GPT-3.5-Turbo without any further training. By performing automatic evaluation, we show that T5 (\textit{\textit{with long prompt}}) outperforms all other models, but still falls short of the human baseline. Under human evaluation criteria, Text-Davinci-003 usually shows better results than other models under various prompt settings. Even in the case of human evaluation criteria, QG models mostly fall short of the human baseline. Our code and dataset are available at: \url{https://github.com/my625/PromptQG}

\end{abstract}






\maketitle

\section{Introduction}

    


The primary objective of the automated question generation task (AQG) is to automatically produce questions based on textual or knowledge data. Prompt-based QG refers to the approach of generating questions using a prompt or stimulus text, which enables providing more information while producing questions \cite{r29, r30}. 

Previous studies in AQG \citep{r14,r15,r18} employ single-hop question-answering (QA) datasets such as SQuAD \cite{r10}, which are representative of QA research, as well as multi-hop QA datasets such as HotpotQA \cite{r11}. But they were considered unsuitable for this particular study, which specifically focuses on educational materials like textbooks for real-life classroom scenarios.

    Our proposed dataset \textit{\textit{EduProbe}} is sourced from NCERT \footnote{\url{https://en.wikipedia.org/wiki/National_Council_of_Educational_Research_and_Training}} textbooks and covers a wide range of subjects and grade levels from $6^{th}$ to $12^{th}$ standard.
    We carefully annotate this dataset as quadruples of 1) \textit{Context}: a segment upon which the question is formed; 2) \textit{Long Prompt}: a long textual cue for the question (\textit{i.e.}, a longer sequence of words or phrases, covering the main theme of the \textit{context}); 3) \textit{Short Prompt}: a short textual cue for the question (\textit{\textit{i.e.}}, a condensed representation of the key information or focus of the \textit{context}). Primarily, we select a noun phrase from the beginning portion of the \textit{context} to serve as a \textit{short prompt}; 4) \textit{Question}: a deep question that aligns with the \textit{context} and is coherent with the \textit{prompts}. After annotation, we gather a total of 3,502 <\textit{Context}, \textit{Long Prompt}, \textit{Short Prompt}, \textit{Question}> quadruples to form the \textit{\textit{EduProbe}} dataset. An example is given in Table \ref{tab:t7}.

\begin{table}[h!]
\caption{ One instance within our \textit{\textit{EduProbe}} dataset.}
\label{tab:t7}
  \centering
  \normalsize
  \renewcommand{\arraystretch}{1.3}   
\scalebox{0.65} {
  \begin{tabular}{|p{0.67\textwidth}|}
\hline
{\textcolor{red}{\bf Context}:} Purchasing power parity (PPP) is an economic indicator that signifies the purchasing power of the currencies of various nations of the world against each other. It helps in comparing living standards between different countries and estimating economic productivity.
\\ 
{\textcolor{Green}{\bf Long Prompt}:} purchasing power parity helps
\\ 
{\textcolor{Green}{\bf Short Prompt}:} purchasing power
\\
{\textcolor{Green}{\bf Gold Standard Question}} : What does purchasing power parity do?
\\
\hline 
\end{tabular}
}

\end{table}
    
   
    To go into the diversity and depth of questions in our created dataset, we classify questions based on the first two words in the question and make comparisons to other frequently used QA datasets, such as SQuAD, and HotpotQA. The corresponding table is provided as Table \ref{tab:t1}. As stated in \cite{r12}, questions phrased with "\textit{Why is}" and "\textit{How do}" are often considered to be deep questions. Table \ref{tab:t1} shows that our proposed dataset, \textit{\textit{EduProbe}} consists of more deep questions (starting with "\textit{Why is}", and "\textit{How do}") compared to SQuAD, and HotpotQA. Following the criteria in \cite{r35}, we further categorize these questions based on their reasoning type. It turns out that 46.14\% of questions in \textit{\textit{EduProbe}} include deep reasoning.

\begin{table}[h!]
   \caption{Leading bigrams that occur most frequently in \textit{\textit{EduProbe}}, SQuAD, and HotpotQA.}
    \label{tab:t1}
    \centering
\scalebox{0.72}{

\begin{tabular}{ |c|c|c|c|c|c|}
    \hline
        \textbf{\textit{\textit{EduProbe}}} & \textbf{\%} & \textbf{SQuAD} & \textbf{\%} & \textbf{HotpotQA} & \textbf{\%}\\
        \hline
          What is & 17.70 & What is & 8.5 & What is & 5.0\\
          What are & 15.10 & What was & 5.3 & Who was & 2.1 \\
          \textbf{Why is} & 7.31 & How many & 4.9 & What was & 2.0 \\
          What was & 2.94 & When did & 3.1 & In what & 1.8 \\
          Which is & 2.79 & In what & 2.9 & When was & 1.7 \\
          How many & 2.34 & What did & 2.8 & Who is & 1.6 \\
          \textbf{How do} & 2.08 & When was & 2.1 & How many & 1.0 \\
          Who was & 1.91 & Who was & 2.1 & In which & 0.9 \\
          Who is & 1.91 & What does & 1.7 & What year & 0.9 \\
          What do & 1.88 & What are & 1.7 & Are both & 0.9\\                  
     \hline
     \end{tabular}
 }

\end{table}


     In this work, we fine-tune four transformer-based large language models (LLMs) such as Pegasus \cite{r6}, T5 \cite{r7},  MBART \cite{r8}, and BART \cite{r9} on our proposed dataset \textit{EduProbe}. With the recent advancement of general-purpose LLMs\footnote{ Details of the two LLMs are available at: \url{https://platform.openai.com/docs/models/}} such as Text-Davinci-003 and GPT-3.5-Turbo, the question arises of whether these general-purpose LLMs can be used for QG without any further training. So we also explore pre-trained LLMs such as Text-Davinci-003 and GPT-3.5-Turbo for prompt-based QG. Through a comprehensive evaluation, we observe that T5 (\textit{\textit{with long prompt}}) outperforms other models in all automated evaluation metrics (see Section \ref{sec7}). In the case of automated metrics, the QG models still fall short of the human baseline. Regarding human evaluation (see Section \ref{sec8}), Text-Davinci-003 (\textit{\textit{with long prompt}}) shows the best proficiency in generating questions in terms of \textit{grammaticality}. Furthermore, Text-Davinci-003 (\textit{\textit{with short prompt}}) demonstrates commendable proficiency in generating questions in terms of \textit{appropriateness}, and \textit{relevance}. On the other hand, Text-Davinci-003 (\textit{\textit{without prompt}}) distinguishes itself by generating questions that are \textit{novel} and \textit{complex}. Although in the case of manual evaluation criteria, the QG models still fall short of the human baseline except for Text-Davinci-003 and GPT-3.5-Turbo under \textit{novelty} and \textit{complexity} criteria.
    
    In summary, the contributions of this paper are as follows: (i) Developing a dataset called \textit{\textit{EduProbe}}, for school-level subjects, namely History, Geography, Economics, Environmental Studies, and Science, which to the best of our knowledge is not present; (ii) We perform an in-depth comparative study of prompt-based (\textit{e.g.}, \textit{long} and \textit{short prompt}) techniques, and \textit{\textit{without prompt}}-based technique utilizing state-of-the-art (SOTA) LLMs on our proposed dataset, \textit{\textit{EduProbe}}. We evaluate them using automated metrics; 
    (iii) Since automated metrics have their own limitations, in terms of evaluating deep questions, we also perform the human evaluation of the generated questions with the help of school-level students and teachers, thereby drawing various meaningful insights.


    The remainder of the paper is organized as follows. We discuss the relevant literature in Section~\ref{sec2}. We present the motivation of our work in Section~\ref{sec13}. We then define the task in Section~\ref{sec3}, and discuss the dataset in Section~\ref{sec4}. We describe the methodology in Section~\ref{s3}, and the experimental setting in Section~\ref{sec6}. We present the automated and human evaluation metrics, results in Section~\ref{sec7}, and Section~\ref{sec8} respectively. We have a general discussion on the analysis of the results in Section~\ref{sec9}. Finally, we conclude our work in Section~\ref{sec10}.
    
\section{Related Work}\label{sec2}
Previous works of QG utilize sequence-to-sequence (Seq2Seq) models \cite{r14,r15}, to produce questions based on various aspects of the sentence, including its focus, type, and specific relationships. 
A model is proposed by \citet{r19} which is made up of four components: a document encoder that encodes the input document, a semantic graph encoder that embeds the document-level semantic graph using a gated graph neural network based on attention, a content selector that identifies important information from the semantic graph suitable for generating questions, and a question decoder that generates questions based on the enhanced document representation. 
\citet{r20} introduce a framework consisting of two main parts: the question generator and QG-specific rewards. The question generator utilizes a Seq2Seq framework with attention, copying, and coverage mechanisms, similar to existing neural QG works. During training, the model learns by maximizing the likelihood of correct questions. However, this basic question generator faces a problem called exposure bias. To address this issue, they introduce three QG-specific rewards to assess the quality of the questions generated by the basic model. These rewards focus on assessing the \textit{fluency}, \textit{relevance}, and \textit{answerability} of the questions. 

 \subsection{Prompt-based QG}
 Current works \cite{r29,r30} explore a few prompt-based techniques for QG.
 \citet{r29} build a large dataset called KHANQ by annotating each data sample as a triple of <\textit{Context}, \textit{Prompt}, \textit{Question}> and explore prompt-based QG with LLMs such as BERT Generation \cite{r31}, BART \cite{r9}, GPT2 \cite{r32}, T5 \cite{r7}, and UniLM \cite{r33}. The prompts used in KHANQ have been designed on the basis of the learner's background knowledge and understanding of the subject.
 Prompt-based fine-tuning is employed by \citet{r30} to create multi-hop questions. The methodology for this task involves a series of tasks starting with QG, followed by QA, which is performed repeatedly in cycles to develop a robust methodology for the QG task. They use T5 to train both the QG and the QA models. Also, question paraphrasing is being performed, which adds to the robustness of the method. Finally, prompt-based fine-tuning is performed to generate quality questions. They generate a prompt by selecting relevant words associated with the correct answer. 
\subsection{Use of LLMs for QG} \label{s1}
Here, we discuss the following LLMs, which we explore for the task of QG in this study: \\
\textbf{Text-Davinci-003} (abbreviated as Davinci) is a model by OpenAI which has 175 billion parameters. During the training process, a combination of supervised and unsupervised learning methods is used. The specific details of the training data sources used for Davinci have not been publicly disclosed by OpenAI. However, training data are known to consist of a diverse range of sources, including web pages, books, scientific articles, and various other forms of human-written text. The maximum input length supported by Davinci is 4,097 tokens.
Davinci is available at: \url{https://platform.openai.com/docs/models/gpt-3-5}.\\
\textbf{GPT-3.5-Turbo} (abbreviated as ChatGPT), an extension of the GPT-3 architecture, which has around 154 billion parameters and underwent training using various text sources such as web pages, books, scientific articles, and more. Its training methods included supervised and reinforcement learning techniques. The primary focus of its optimization was to improve speed, performance, and resource efficiency. It has a maximum input token limit of 4,096. ChatGPT is available at: \url{https://platform.openai.com/docs/models/gpt-3-5}.\\
\textbf{T5} \cite{r7} is based on the transformer architecture and trained using a large-scale dataset consisting of diverse text sources. T5-large has a huge number of parameters, specifically around 770 million, enabling it to capture intricate patterns and relationships in text data. The maximum input length supported by T5-large is 512 tokens. The extensive pre-training and fine-tuning process of T5 makes it a powerful tool for generating high-quality questions and producing accurate natural language output. Here, we use \texttt{T5-large} (\url{https://huggingface.co/t5-large}) for the QG task.\\
\textbf{Pegasus} \cite{r6} is a transformer-based model designed for text summarization, using an encoder-decoder architecture with self-attention mechanisms to capture long-range dependencies. With approximately 568 million parameters, Pegasus-large can handle complex summarization tasks and generate high-quality summaries. Its input limit is 1,024 tokens. It has the potential to be utilized for the QG task. Here, we utilize \texttt{Pegasus-large} 
 (\url{https://huggingface.co/google/pegasus-large}) for the purpose of QG.\\
\textbf{MBART} \cite{r8} (Multilingual Bidirectional Auto-Regressive Transformers) incorporates a transformer-based architecture, which allows it to effectively capture contextual information and generate high-quality translations. It leverages pre-training on a large multilingual corpus to learn cross-lingual representations. In terms of the number of parameters, MBART-large-50 has around 610 million parameters. It has a maximum input token limit of 1,024. It can be utilized for the QG task. Here, we utilize \texttt{MBART-large} (\url{https://huggingface.co/facebook/MBART-large-50}) for QG task.\\
\textbf{BART} \cite{r9}(Bidirectional Auto-Regressive Transformers) comprises a bidirectional encoder and a left-to-right decoder. During pre-training, it shuffles the order of sentences and uses a unique method of infilling where sections of text are replaced with a mask token. BART-large has around 406 million parameters. The maximum input length supported by BART is 1,024 tokens. It can be used for QG tasks. Here, we use \texttt{BART-large} (\url{https://huggingface.co/facebook/BART-large}) for the QG task.
\subsection{Datasets used for QG}
Due to the data-centric nature of QG, the QG methods mentioned above leverage the availability of large-scale QA datasets, such as SQuAD, HotpotQA, TriviaQA \citep{r39}, Natural Questions corpus \citep{r38}, QuAC \citep{r40}, OpenBookQA \citep{r41}, etc. According to \citet{r35}, these corpora are limited to the generation of simple fact-based questions. Furthermore, as stated in \cite{r26, r37}, the majority of these QA datasets are borrowed or crowd-sourced from open-source platforms such as Wikipedia articles, and the questions generally do not incorporate multiple sentences as their basis. There is a notable QG dataset for educational purposes called LearningQ \cite{r25}, which utilizes complete articles or videos as contexts, resulting in a substantial portion of sentences within the contexts being irrelevant to the specific target question. In contrast, we utilize explanatory answers that contain comprehensive knowledge points relevant to the question.

\section{Motivation}\label{sec13}
 The creation of high-quality questions is a fundamental task for educators seeking to foster deep understanding and critical thinking in students. However, the process of designing educational questions manually is often burdensome and time-consuming \cite{r29}. Through this research effort, our aim is to provide a valuable tool that empowers educators to create descriptive and reasoning-based questions more efficiently. By allowing teachers to allocate more time to classroom interactions and student participation, our proposed QG method has the potential to positively impact teaching practices and enhance learning outcomes.

Previous research in QG \citep{r14,r15,r18} predominantly emphasizes the generation of fact-based questions that relate to a single piece of information derived from a single sentence. Moreover, current QA datasets such as SQuAD, HotpotQA, TriviaQA, QuAC, OpenBookQA, etc. do not align with the requirements of generating school-level educational questions since they do not have deep questions that are reasoning-based and descriptive in nature.

So, we curate a new dataset called \textit{EduProbe} to tackle the complexities of school-level educational QG. The proposed dataset serves as the foundation for our investigation into the efficacy of prompt-based methods for QG, which involves providing the system with explicit hints or triggers to generate deep questions. The prompt-based AQG process offered by our approach not only reduces the time required for question development, but also improves the overall quality and variety of questions generated.

Our work has the following three key differences from previous works on QG: (i) Our created dataset \textit{\textit{EduProbe}} is geared towards creating questions that are more educationally oriented in the context of school-level subjects; (ii) We explore different types of prompt-based techniques (\textit{e.g.}, \textit{long prompt}, \textit{short prompt}, and \textit{\textit{without prompt}}) with SOTA LLMs (\textit{e.g.}, Text-Davinci-003, GPT-3.5-Turbo, etc.) to provide the QG models additional guidance on what information to emphasize more when generating questions that still have not been explored earlier in a detailed manner; (iii) Our proposed prompt-based approaches are capable of generating a wide variety of questions from a single context which has not been explored in previous works.

\section{Task Definition}\label{sec3}
\begin{figure*}
  \centering
  \includegraphics[width=\linewidth]{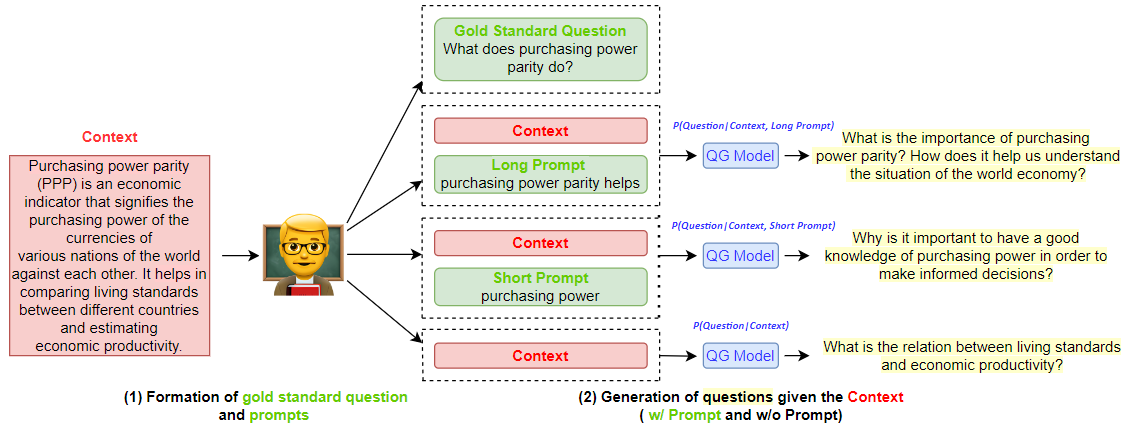}
  \caption{A diagrammatic representation of the pipeline process utilized to generate questions.} \label{fig4}
\end{figure*}

In this section, we define three different prompt settings explored in our study:
\begin{itemize}

  \item \textbf{With Long Prompt}: Given a dataset $D$, where each data point $d \in D$ is represented as a 4-tuple <\textit{Context}, \textit{Long Prompt}, \textit{Short Prompt}, \textit{Question}>, our task is to learn a probabilistic model \textit{P(Question | Context, Long Prompt)} that can generate a relevant question in the context of the given information.
   
  \item \textbf{With Short Prompt}: Given a dataset $D$, where each data point $d \in D$ is represented as a 4-tuple <\textit{Context}, \textit{Long Prompt}, \textit{Short Prompt}, \textit{Question}>, our task is to learn a probabilistic model \textit{P(Question | Context, Short Prompt)} that can generate a relevant question in the context of the given information.
  
  \item \textbf{Without Prompt}: Given a dataset $D$, where each data point $d \in D$ is represented as a 4-tuple <\textit{Context}, \textit{Long Prompt}, \textit{Short Prompt}, \textit{Question}>, our task is to learn a probabilistic model \textit{P(Question | Context)} that can generate a relevant question in the context of the given information.
\end{itemize}
Figure \ref{fig4} illustrates the schematic of the pipeline process used for QG.

\section{Dataset}\label{sec4}
\textbf{Data Collection and Annotation:} There is no public dataset available to conduct our experiments on prompt-based QG in an educational setting. Therefore, we produce a dataset called \textit{\textit{EduProbe}} by manually creating \textit{question-answer} pairs from segments of varying lengths taken from a diverse set of chapters present in the National Council of Educational Research and Training (NCERT) textbooks on History, Geography,  Economics, Environmental Studies, and Science from the $6^{th}$ standard to the $12^{th}$ standard. 

Firstly, the annotators are instructed to go through the selected chapters line by line and generate \textit{ question-answer} pairs by considering only the relevant portions of the information from the NCERT textbooks. To establish the \textit{context}, we instruct the annotators to review the generated answers. Upon analysis, we observe that the majority of the answers consist of comprehensive explanations related to the questions' pertinent knowledge points. Hence, these answers prove suitable for serving as the \textit{context} for the question. Secondly, we instruct the annotators to select a sequence of words or phrases which cover the main theme of the \textit{context} to serve as a \textit{long prompt} for the question. Lastly, annotators are asked to pick up a noun phrase from the beginning portion of the \textit{context} to act as a \textit{short prompt}. In this manner, each \textit{question-answer} pair in \textit{\textit{EduProbe}} is carefully annotated as a quadruple of <\textit{Context}, \textit{Long Prompt}, \textit{Short Prompt}, \textit{Question}>.



Two annotators (two graduate students) with adequate subject knowledge and experience were involved in manually annotating the data samples.\\
\\
\textbf{Data Statistics:}  We carefully curate 3,502 question-answer pairs, of which 858 pairs are related to History, 861 pairs are related to Geography, 802 pairs are related to Economics, 606 pairs are related to Environmental Studies, and 375 pairs are related to Science. On average, the length of the \textit{context}, \textit{long prompt}, \textit{short prompt}, and question are 55.27 words, 6.80 words, 2.15 words, and 7.16 words, respectively. 

Comparing the KHANQ dataset and our proposed dataset \textit{\textit{EduProbe}} we observe that the average length of the \textit{long prompt} and \textit{short prompt} in our dataset is 6.80 and 2.15 words, respectively. But in the KHANQ dataset, the average \textit{prompt} length is 14.12 words. However, the KHANQ dataset is not publicly available for research purposes. 
\\
\\
\textbf{Question Types:} In order to gain a deeper understanding of the question attributes, we conduct a detailed manual analysis of a subset of 65 distinct questions randomly sampled from the \textit{\textit{EduProbe}} dataset. These questions were categorized according to the criteria specified in \cite{r35}. Here, we present a summary of the most commonly found question types in \textit{\textit{EduProbe}} and their corresponding examples.

\begin{itemize}
    \item \textbf{Procedural Questions}: Out of the questions we examine, 18.46\% of them are about the procedures or methods used to achieve a specific outcome. Most of these questions begin with “\textit{How}” and are followed by a modal verb, an auxiliary verb, or “\textit{to}”.
    
     \begin{itemize}
         \item \textit{How did Pandita Ramabai break stereotypes?}
         \item \textit{How did Brahmo Samaj reform Indian society?}
     \end{itemize}

    \item \textbf{Cause Questions}: We observe that 15.38\% of the questions examined are focused on the reason or cause behind a concept or event. Most of these questions began with the word “\textit{Why}” and were subsequently followed by a modal verb, an auxiliary verb, or their negative counterparts.
    \begin{itemize}
        \item \textit{Why is the Ganges river dolphin blind?}
        \item \textit{Why is urban waste disposal a serious problem in India?}
    \end{itemize}

    \item \textbf{Verification Questions}: We discover that 9.23\% of the examined questions are concerned with verifying the trustworthiness of a concept or event. Most of these questions are formulated as general questions that originate with verbs, modal verbs, or auxiliary verbs.
    \begin{itemize}
        \item \textit{Does universal basic income (UBI) reduce poverty?}
         \item \textit{Are Vedas older than Puranas?}
    \end{itemize}
    
    \item \textbf{Consequence Questions}: We find that 12.30\% of the questions analyzed focused on the consequences or outcomes resulting from a particular event. Most of these questions used phrases such as “\textit{What happens}”, “\textit{How does it affect}”, etc.
    \begin{itemize}
        \item \textit{What happens if oceans acidify?}
         \item \textit{How does the government deficit affect the economy?}
    \end{itemize}
        
\end{itemize}

According to the findings of \cite{r35}, there are four categories of questions that require profound reasoning, namely cause, consequence, judgemental, and procedural. Relating these categories to \textit{\textit{EduProbe}}, we observe that the procedural questions, the cause questions, and the consequence questions are three specific categories that involve deep reasoning, representing 46.14\%.

\section{Methodology} \label{s3}

We experiment with various prompt-based settings with LLMs. There are two main categories of LLMs explored in our study: (1) Pre-trained General-purpose LLMs; (2) Fine-tuned Domain-specific LLMs.
\subsection{Pre-trained General-purpose LLMs}
We try Text-Davinci-003 (abbreviated as Davinci) and GPT-3.5-Turbo (abbreviated as ChatGPT) model using OpenAI API\footnote{\url{https://platform.openai.com/docs/api-reference/completions}}. For the pre-trained general-purpose LLMs, we have to pass a \textit{prompt} as input to the LLM which will consist of the instruction for the task of QG, along with the \textit{context} and \textit{prompt} (if needed). The LLM will generate text based on the given prompt, which will be our desired output.
For each model, we try three different variations based on prompts that are as follows:\\
a) \textbf{With Long Prompt}: The prompt we use is ``\textit{Given the context  <\textit{\texttt{Context}}> and the long prompt <\texttt{Long Prompt}>}, \textit{generate a \texttt{Question}}''.\\
b) \textbf{With Short Prompt}: The prompt we apply is ``\textit{Given the context <\textit{\texttt{Context}}> and the short prompt <\texttt{Short Prompt}>}, \textit{generate a \texttt{Question}}''. \\
c) \textbf{Without Prompt}: The prompt we utilize is ``\textit{Given the context <\texttt{Context}>}, \textit{generate a \texttt{Question}}''.

\subsection{Fine-tuned Domain-specific LLMs}

\begin{figure}[h!]
  \centering
  \includegraphics[width=0.90\linewidth]{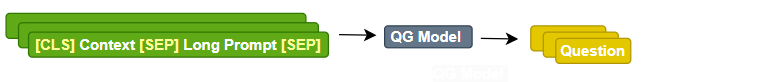}
  \caption{With Long Prompt QG Model.} \label{fig1}
\end{figure}

\begin{figure}[h!]
  \centering
  \includegraphics[width=0.90\linewidth]{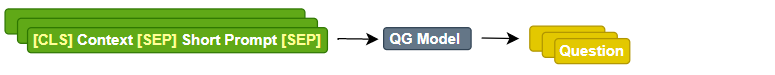}
  \caption{With Short Prompt QG Model.} \label{fig2}
\end{figure}

\begin{figure}[h!]
  \centering
  \includegraphics[width=0.90\linewidth]{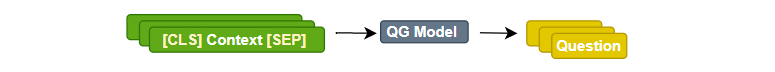}
  \caption{Without Prompt QG Model.} \label{fig3}
\end{figure}

We also fine-tune four pre-trained transformer-based QG models (or LLMs), namely Pegasus \cite{r6}, T5 \cite{r7}, MBART \cite{r8}, and BART \cite{r9} obtained from the open-source library\footnote{\url{https://huggingface.co/models}} named Huggingface, in our proposed dataset \textit{\textit{EduProbe}}. For every model, we try the three different prompt-based techniques which are as follows:\\
a) \textbf{With Long Prompt}: During fine-tuning, we format the input sequence as:( \textit{[\texttt{CLS}] \texttt{Context} [\texttt{SEP}] \texttt{Long Prompt} [\texttt{SEP}]}, \textit{[\texttt{CLS}] \texttt{Question} [\texttt{SEP}]}) pair. We provide \textit{[\texttt{CLS}] \texttt{Context} [\texttt{SEP}] \texttt{Long Prompt} [\texttt{SEP}]} as input to predict the \texttt{Question} during test time, as shown in Figure \ref{fig1}.\\ 
 \\
b) \textbf{With Short Prompt}: During fine-tuning, we format the input sequence as:( \textit{[\texttt{CLS}] \texttt{Context} [\texttt{SEP}] \texttt{Short Prompt} [\texttt{SEP}]}, \textit{[\texttt{CLS}] \texttt{Question} [\texttt{SEP}]}) pair. We provide \textit{[\texttt{CLS}] \texttt{Context} [\texttt{SEP}] \texttt{Short Prompt} [\texttt{SEP}]} as input to predict the \texttt{Question} during test time, as shown in Figure \ref{fig2}.\\
\\
c) \textbf{Without Prompt}: During fine-tuning, we format the input sequence as:( \textit{[\texttt{CLS}] \texttt{Context} [\texttt{SEP}], [\texttt{CLS}] \texttt{Question} [\texttt{SEP}]}) pair. We provide \textit{[\texttt{CLS}] \texttt{Context} [\texttt{SEP}]} as input to predict the \texttt{Question} during test time, as shown in Figure \ref{fig3}.\\

\section{Experimental Settings}\label{sec6}
In our experiment, we randomly sample 80\% of the data in \textit{EduProbe} for training and the rest for testing. The experiments are run on an NVIDIA Tesla P100 16GB GPU and the models are optimized using the Adam optimizer \cite{r23}. The specific hyperparameter configurations of the LLMs used in our experiments are given in Table \ref{tab:t5}.

\begin{table}[h!]
    \caption{Hyperparameters of the LLMs used in our work.}
     \label{tab:t5}
    \centering
\scalebox{0.71}{

    \begin{tabular}{ |c|c| }
    \hline
         \textbf{Model} & \textbf{Hyperparameters} \\
         \hline
         Davinci & max tokens: 50, presence penalty: 1.0, frequency penalty: 0.0, temperature: 0.7\\
         ChatGPT & max tokens: 50, temperature: 0.7\\\hline
         Pegasus & learning rate: 2e-3, epochs: 6, batch size: 1, input length: 512 \\
         T5 & learning rate: 2e-3, epochs: 6, batch size: 2, input length: 512\\
         MBART & learning rate: 1e-3, epochs: 8, batch size: 1, input length: 512\\
         BART & learning rate: 1e-3, epochs: 8, batch size: 1, input length: 512\\

    \hline
   
    \end{tabular}
  }  

\end{table}

\section{Automatic Evaluation}\label{sec7}
In this section, we present the main results in \textit{EduProbe}, using the methodology explained in Section \ref{s3}.
\subsection{Evaluation Metrics} \label{s2}
We use the following popular metrics that compare a QG model-generated question with the gold standard question: \\
\textbf{ Rouge} \cite{r3} (Recall-Oriented Understudy for Gisting Evaluation) is widely utilized as a metric to assess the quality of generated summaries by summarization models. Here, we compute Rouge-2 precision, Rouge-2 recall, and Rouge-2 F1 score to evaluate the bigram overlap between the QG model-generated questions and the reference gold standard questions. Furthermore, Rouge-L precision, Rouge-L recall, and Rouge-L F1 scores are calculated to measure the longest common subsequence-based match between the generated questions and the gold standard questions.\\
\textbf{Meteor}\cite{r1} (Metric for Evaluation of Translation with Explicit ORdering) calculates the harmonic mean of unigram precision and recall, which is commonly used to evaluate machine translation results.  In our case, we apply this metric to measure the unigram overlap between a QG model-generated question and the reference gold standard question. \\
\textbf{ CHrF}\cite{r5} (Character n-gram F-score) is a metric that evaluates the similarity between the generated output and the reference summaries at the character level. Here, CHrF calculates the F-score based on the precision and recall of the matching character n-grams between the QG model-generated questions and the reference gold-standard questions.\\
\textbf{BLEU}\cite{r2}(Bilingual Evaluation Understudy) is a widely used metric to evaluate the quality of a machine-generated text. Here, we calculate the overlap between the QG model-generated question and the reference gold question based on n-gram matches. BLEU calculates a score ranging from 0 to 1, with higher scores indicating better quality.\\
\textbf{BERTScore}\cite{r4} is a metric that we utilize to measure the similarity between the generated question and the reference gold question using contextual embeddings from a pre-trained BERT model. It computes a score based on the cosine similarity of the embeddings, capturing both lexical and contextual similarities.

We utilize the implementations of the aforementioned metrics from the SummEval package \footnote{\url{https://github.com/Yale-LILY/SummEval}}.

In order to compare with human performance, we appoint two high school teachers who were not involved in the annotation process and request them to undertake the same task as the models, generating questions based on the given \textit{context} and \textit{prompt} settings for all data samples from the test set. To minimize subjective variations, they were instructed to collaborate and reach a consensus while formulating their responses.

\subsection{Results}
We present the results of automated evaluation metrics for different models, in order to investigate the influence of prompts. Table \ref{tab:t3} represents the results of automatic evaluation of LLMs under different prompt settings for QG. T5 performs the best across all metrics in the \textit{long prompt} setting. When the \textit{prompt is short}, T5 achieves the highest scores in ROUGE-2 precision and ROUGE-L recall. BART obtains the best results in ROUGE-2 Recall, ROUGE-2 F1, ROUGE-L Precision,  ROUGE-L F1, METEOR, CHrF, BLEU, and BERTScore under the \textit{short prompt} setting. Furthermore, T5 outperforms the other models in terms of all metrics, except for BERTScore, in the \textit{without prompt} setting and BART achieves the highest BERTScore.

Compared to other models, T5 and BART exhibit superior performance in automatic evaluation across various prompt settings (\textit{e.g.}, \textit{long prompt}, \textit{short prompt}, and \textit{\textit{without prompt}}). However, human references consistently achieve the highest scores across all automated metrics and prompt settings. This observation emphasizes the fact that SOTA LLMs still have not reached the level of human performance on our \textit{\textit{EduProbe}} dataset.

\begin{table*}
\caption{Automatic evaluation results for different LLMs in \textit{\textit{EduProbe}} with ROUGE2-Precision, ROUGE2-Recall, ROUGE2-F1, ROUGEL-Precision, ROUGEL-Recall, ROUGEL-F1, METEOR, CHrF, BLEU, and BERTScore. The highest value for any metric in \textit{long prompt}, \textit{short prompt}, and \textit{\textit{without prompt}} setting achieved by any model is shown in \textcolor{blue}{blue}. The highest value for any metric achieved by any model is \underline{\textcolor{blue}{underlined}}. }
\label{tab:t3}    
    \centering
\scalebox{0.72}{
\begin{tabular}{|c|c|c|c|c|c|c|c|c|c|c|}
\hline
\textbf{Model} & \textbf{\begin{tabular}[c]{@{}c@{}}ROUGE-2 \\ Precision\\ \end{tabular}} & \textbf{\begin{tabular}[c]{@{}c@{}}ROUGE-2 \\ Recall\\ \end{tabular}} & \textbf{\begin{tabular}[c]{@{}c@{}}ROUGE-2 \\ F1\\ \end{tabular}} & \textbf{\begin{tabular}[c]{@{}c@{}}ROUGE-L\\ Precision\\ \end{tabular}} & \textbf{\begin{tabular}[c]{@{}c@{}}ROUGE-L\\ Recall\\ \end{tabular}} & \textbf{\begin{tabular}[c]{@{}c@{}}ROUGE-L \\ F1 \\ \end{tabular}} & \textbf{METEOR} & \textbf{\begin{tabular}[c]{@{}c@{}} CHrF \\ (\%) \\ \end{tabular}} & 
\textbf{\begin{tabular}[c]{@{}c@{}}BLEU \\ (\%)\\ \end{tabular}}& 
\textbf{BERTScore} \\ \hline



\multicolumn{11}{|c|}{\textit{\textbf{\textcolor{Green}{\textit{With Long Prompt}}}}}         \\ \hline
\textbf{Human Baseline}  & \textbf{0.517} &  \textbf{0.843} & \textbf{0.626} & \textbf{0.588} & \textbf{0.891} & \textbf{0.695} & \textbf{0.531} & \textbf{76.30} & \textbf{46.57} & \textbf{0.860}\\ \hline
\multicolumn{11}{|c|}{\textbf{Pre-trained General-purpose LLMs}} \\\hline
Davinci & 0.409 & 0.726 & 0.491 &  0.499 & 0.812 & 0.603 & 0.443 & 68.23 & 23.24 & 0.803\\

ChatGPT & 0.391 & 0.706 & 0.476 &  0.484 & 0.793 & 0.592 & 0.423 & 66.36 & 22.44 & 0.790\\ \hline
\multicolumn{11}{|c|}{\textbf{Fine-tuned Domain-specific LLMs}} \\\hline
Pegasus & 0.329 &  0.798 & 0.453 & 0.413 & 0.885 & 0.552 & 0.411 & 67.95 & 27.78 &0.770 \\ 

T5   & \textbf{\underline{\textcolor{blue}{0.483}}}  &  \textbf{\underline{\textcolor{blue}{0.800}}} & \textbf{\underline{\textcolor{blue}{0.575}}} & \textbf{\underline{\textcolor{blue}{0.566}}} & \textbf{\underline{\textcolor{blue}{0.888}}} & \textbf{\underline{\textcolor{blue}{0.668}}} & \textbf{\underline{\textcolor{blue}{0.503}}} & \textbf{\underline{\textcolor{blue}{74.40}}} & \textbf{\underline{\textcolor{blue}{42.97}}} & \textbf{\underline{\textcolor{blue}{0.818}}}   \\ 

MBART & 0.424  & 0.750 & 0.526 & 0.527 & 0.860 & 0.640 & 0.417 & 71.05 & 33.55 & 0.786 \\ 
BART   &  0.460 & 0.794 & 0.573 & 0.548 & 0.887 & 0.666 & 0.443 & 73.72 & 36.47 & 0.809 \\ \hline

\multicolumn{11}{|c|}{\textit{\textbf{\textcolor{Green}{\textit{With Short Prompt}}}}}         \\ \hline
\textbf{Human Baseline} & \textbf{0.324} &  \textbf{0.579} & \textbf{0.418} & \textbf{0.468} & \textbf{0.758} & \textbf{0.563} & \textbf{0.377} & \textbf{61.52} & \textbf{24.92} & \textbf{0.778}\\ \hline
\multicolumn{11}{|c|}{\textbf{Pre-trained General-purpose LLMs}} \\\hline
Davinci & 0.263 & 0.492 & 0.319 &  0.429 & 0.723 & 0.529 & 0.313 & 55.62 & 20.86 & 0.739\\

ChatGPT & 0.260 & 0.486 & 0.304 &  0.418 & 0.720 & 0.522 & 0.309 & 54.83 & 19.54 & 0.720\\\hline

\multicolumn{11}{|c|}{\textbf{Fine-tuned Domain-specific LLMs}} \\\hline
Pegasus & 0.240 &  0.496 & 0.312 & 0.374 & 0.708 & 0.477 & 0.331 & 54.41 & 18.54 & 0.723 \\ 

T5   & \textbf{\textcolor{blue}{0.301}} & 0.530 & 0.368 & 0.448 & \textbf{\textcolor{blue}{0.742}} & 0.542 & 0.341 & 58.76 & 21.15 & 0.742   \\ 

MBART & 0.237  & 0.464  & 0.308 & 0.385 & 0.680 & 0.483 & 0.307 & 53.56 & 17.64 & 0.718 \\ 
BART   &  0.300 & \textbf{\textcolor{blue}{0.541}} & \textbf{\textcolor{blue}{0.377}} & \textbf{\textcolor{blue}{0.449}} & 0.740 & \textbf{\textcolor{blue}{0.549}} & \textbf{\textcolor{blue}{0.346}} & \textbf{\textcolor{blue}{59.52}} & \textbf{\textcolor{blue}{21.82}} & \textbf{\textcolor{blue}{0.756}} \\ \hline

\multicolumn{11}{|c|}{\textit{\textbf{\textcolor{Green}{\textit{Without Prompt}}}}}       \\ \hline

\textbf{Human Baseline} & \textbf{0.323} &  \textbf{0.532} & \textbf{0.390} & \textbf{0.466} & \textbf{0.723} & \textbf{0.553} & \textbf{0.355} & \textbf{58.23} & \textbf{23.49} & \textbf{0.758}\\ \hline
\multicolumn{11}{|c|}{\textbf{Pre-trained General-purpose LLMs}} \\\hline
Davinci & 0.273 & 0.462 & 0.327 &  0.413 & 0.666 & 0.506 & 0.283 & 54.86 & 20.44 & 0.729\\

ChatGPT & 0.266 & 0.442 & 0.319 &  0.401 & 0.645 & 0.492 & 0.266 & 52.83 & 19.44 & 0.710\\ \hline
\multicolumn{11}{|c|}{\textbf{Fine-tuned Domain-specific LLMs}} \\\hline
Pegasus & 0.214 &  0.465 & 0.280 & 0.346 & 0.693 & 0.449 & 0.307 & 51.48 & 16.24 & 0.702 \\ 

T5  & \textbf{\textcolor{blue}{0.306}} & \textbf{\textcolor{blue}{0.501}} & \textbf{\textcolor{blue}{0.368}} & \textbf{\textcolor{blue}{0.455}} & \textbf{\textcolor{blue}{0.706}} & \textbf{\textcolor{blue}{0.539}} & \textbf{\textcolor{blue}{0.322}} & \textbf{\textcolor{blue}{57.03}} & \textbf{\textcolor{blue}{21.59}} & 0.718  \\ 

MBART & 0.219  & 0.414 & 0.281 & 0.373 & 0.642 & 0.464 & 0.293 & 50.46 & 17.34 & 0.706 \\ 
BART   &  0.275 & 0.477 & 0.341 & 0.425 & 0.688 & 0.516 & 0.319 & 54.96 & 20.05 & \textbf{\textcolor{blue}{0.742}} \\ \hline
    \end{tabular}
}

\end{table*}

\section{Human Evaluation}\label{sec8}
Considering the limitations associated with automated metrics in the field of text generation research \cite{r44,r45,r46}, we also conduct a human evaluation by appointing two high school teachers and three high school students who were not engaged in the annotation process and the generation of human baseline questions. Every human evaluator was asked to rate a total of 1,800 questions, taking into account six models and three prompt settings. The rating scale used ranged from 1 (worst) to 5 (best) based on five criteria: 
\textbf{Grammaticality}, which measures the grammatical correctness of the generated question, regardless of the context or prompt; 
\textbf{Appropriateness}, which examines the semantic correctness of the question irrespective of the context or prompt; 
\textbf{Relevance}, which measures the degree to which the generated question is pertinent and aligned with the given context or prompt;
\textbf{Complexity}, which estimates the level of reasoning or cognitive effort required to answer the generated question;
\textbf{Novelty}, which measures the originality and distinctiveness of the generated question in comparison to the gold standard question for the given context.\\
\textbf{Model-wise Evaluation:}
We report the human evaluation results for different models under different prompt settings on the \textit{\textit{EduProbe}} dataset in Table \ref{tab:t4}. Davinci demonstrates superior performance compared to other models in human evaluation in different prompt settings (\textit{e.g.}, \textit{long prompt}, \textit{short prompt}, and \textit{\textit{without prompt}}) in generating questions that exhibit impressive \textit{grammaticality}, \textit{appropriateness}, \textit{relevance}, \textit{complexity}, and \textit{novelty}. However, human references achieve the highest scores on most human criteria, except for the \textit{novelty} under the \textit{long prompt} setting and \textit{complexity} in the \textit{\textit{without prompt}} setting. This also suggests that SOTA LLMs still fall short of reaching the level of human performance in most cases. Although Davinci and ChatGPT overtake human-level performance in terms of producing \textit{complex} questions. \\
\textbf{Inter-annotator Agreement:}
In order to assess the level of agreement among the five annotators assigned to each generated question, we use Fleiss's kappa as a metric for inter-annotator agreement. Our calculations yield agreement scores of 0.49, 0.43, 0.44, 0.39, and 0.32 for \textit{grammaticality}, \textit{appropriateness}, \textit{relevance}, \textit{complexity}, and \textit{novelty}, respectively.
The kappa values in \textit{grammaticality},
\textit{appropriateness}, \textit{relevance} indicate a moderate agreement \cite{r36}, while the kappa results for \textit{complexity} and \textit{novelty} indicate a fair level of agreement.

\begin{table*}[h!]
\caption{Human evaluation results for different LLMs in \textit{\textit{EduProbe}} on grammaticality, appropriateness, relevance, complexity, and novelty. The highest value for any metric in the \textit{long prompt}, \textit{short prompt}, and \textit{\textit{without prompt}} setting achieved by any model is shown in \textcolor{blue}{blue}. The highest value for any metric achieved by any model is \underline{\textcolor{blue}{underlined}}.}
    \label{tab:t4}
    \centering
\scalebox{0.70}{
\begin{tabular}{|c|c|c|c|c|c|}
\hline
\textbf{Model} & \textbf{Grammaticality } & \textbf{Appropriateness} & \textbf{Relevance} & \textbf{Complexity} &  \textbf{Novelty} \\ \hline


\multicolumn{6}{|c|}{\textit{\textbf{\textcolor{Green}{\textit{With Long Prompt}}}}}  \\ \hline
\textbf{Human Baseline} & \textbf{4.95} & \textbf{4.97} & \textbf{4.48} & \textbf{3.98} & \textbf{3.10} \\ \hline
\multicolumn{6}{|c|}{\textbf{Pre-trained General-purpose LLMs}} \\\hline
Davinci & \textbf{\underline{\textcolor{blue}{4.91}}} & \textbf{\textcolor{blue}{4.70}} & \textbf{\textcolor{blue}{4.26}} & \textbf{\textcolor{blue}{3.97}} & \textbf{\textcolor{blue}{3.73}} \\ 
ChatGPT & 4.83 & 4.51 & 4.20 & 3.94 & 3.56 \\ \hline
\multicolumn{6}{|c|}{\textbf{Fine-tuned Domain-specific LLMs}} \\\hline
Pegasus & 4.48 & 4.37 & 3.74 & 3.84 & 3.14 \\ 
T5 & 4.14 &  4.02 & 3.93 & 3.53 & 2.93\\ 
MBART & 3.94 & 3.84 & 3.54 & 3.42 & 3.28 \\ 
BART & 3.97 &  4.00 & 3.90 & 3.76 & 3.20\\ \hline

\multicolumn{6}{|c|}{\textit{\textbf{\textcolor{Green}{\textit{With Short Prompt}}}}}         \\ \hline
\textbf{Human Baseline} & \textbf{4.94} & \textbf{4.95} & \textbf{4.43} & \textbf{4.16} & \textbf{4.08} \\ \hline
\multicolumn{6}{|c|}{\textbf{Pre-trained General-purpose LLMs}} \\\hline
Davinci & \textbf{\textcolor{blue}{4.89}} & \textbf{\underline{\textcolor{blue}{4.80}}} & \textbf{\underline{\textcolor{blue}{4.31}}} & \textbf{\textcolor{blue}{4.12}} & \textbf{\textcolor{blue}{3.93}} \\ 
ChatGPT & 4.85 & 4.76 & 4.28 & 4.03 & 3.81 \\ \hline
\multicolumn{6}{|c|}{\textbf{Fine-tuned Domain-specific LLMs}} \\\hline

Pegasus & 4.50 & 4.48 & 4.06 & 3.81 & 3.44 \\ 
T5 & 4.80 & 4.62 & 4.20 & 3.96 & 3.67\\ 
MBART & 3.38 & 3.28 & 3.08 & 3.14 & 3.17 \\ 
BART & 3.78 & 3.85 & 3.58 & 3.69 & 3.38\\ \hline

\multicolumn{6}{|c|}{\textit{\textbf{\textcolor{Green}{\textit{Without Prompt}}}}}        \\ \hline
\textbf{Human Baseline} & \textbf{4.93} & \textbf{4.96} & \textbf{4.29} & \textbf{4.13} & \textbf{4.25} \\ \hline
\multicolumn{6}{|c|}{\textbf{Pre-trained General-purpose LLMs}} \\\hline
Davinci & \textbf{\textcolor{blue}{4.90}} & \textbf{\textcolor{blue}{4.61}} & \textbf{\textcolor{blue}{4.18}} & \textbf{\underline{\textcolor{blue}{4.25}}} & \textbf{\underline{\textcolor{blue}{4.05}}} \\ 
ChatGPT & 4.82 & 4.46 & 4.15 & 4.18 & 3.96 \\\hline

\multicolumn{6}{|c|}{\textbf{Fine-tuned Domain-specific LLMs}} \\\hline
Pegasus & 4.42 & 4.33 & 3.73 & 3.62 & 3.77 \\ 
T5 & 4.65 & 4.45 & 3.92 & 3.57 & 3.84\\ 
MBART & 3.22 & 3.00 & 2.77 & 3.01 & 3.40 \\ 
BART & 3.81 &  3.98 & 3.60 & 3.60 & 3.72\\ \hline
    \end{tabular}
}

\end{table*}

\begin{table}
\caption{A data sample related to Economics in our \textit{EduProbe} dataset, along with corresponding questions generated by different LLMs using various prompt variations.}
\label{tab:t_2}
  \footnotesize
  \begin{tabular}{|p{0.46\textwidth}|}
\hline
\hline
{\textcolor{Red}{\bf Context}:} Purchasing power parity (PPP) is an economic indicator that signifies the purchasing power of the currencies of various nations of the world against each other. It helps in comparing living standards between different countries and estimating economic productivity.
\\ 
{\textcolor{Green}{\bf Long Prompt}:} purchasing power parity helps
\\ 
{\textcolor{Green}{\bf Short Prompt}:} purchasing power
\\
{\textcolor{Green}{\bf Gold Standard Question}} : What does purchasing power parity do?
\\ 
\hline
{\textcolor{blue}{\bf Davinci (\textit{\textit{with long prompt}})}:} How can the concept of purchasing power parity (PPP) be leveraged to not only compare living standards and estimate economic productivity between different countries but also address global income inequality and promote equitable economic development on a global scale?\\
{\textcolor{blue}{\bf Davinci (\textit{\textit{with short prompt}})}:} How can the concept of purchasing power be harnessed to foster international collaboration and create a more equitable global economic system that promotes shared prosperity and uplifts disadvantaged nations?\\
{\textcolor{blue}{\bf Davinci (\textit{\textit{without prompt}})}:} In what ways can the utilization of purchasing power parity (PPP) as an economic indicator be expanded beyond its traditional applications to promote sustainable development, address socio-economic disparities, and foster international cooperation for the betterment of global economic systems?\\
\hline
{\textcolor{blue}{\bf ChatGPT (\textit{\textit{with long prompt}})}:} What does purchasing power parity (PPP) indicate and how does it assist in comparing living standards and estimating economic productivity among different nations?\\
{\textcolor{blue}{\bf ChatGPT (\textit{\textit{with short prompt}})}:} What does purchasing power parity (PPP) represent and how does it facilitate the comparison of living standards and the estimation of economic productivity across different nations?\\
{\textcolor{blue}{\bf ChatGPT (\textit{\textit{without prompt}})}:} How does purchasing power parity (PPP) serve as an economic indicator to measure the purchasing power of currencies across different nations, and what are its implications for comparing living standards and estimating economic productivity?\\
\hline
{\textcolor{blue}{\bf Pegasus (\textit{With long prompt})}:} What is the importance of purchasing power parity? How does it help us understand the situation of the world economy?
\\
{\textcolor{blue}{\bf Pegasus (\textit{With short prompt})}:} Why is it important to have a good knowledge of purchasing power in order to make informed decisions?
\\
{\textcolor{blue}{\bf Pegasus (\textit{Without prompt})}:} What is the relation between living standards and economic productivity?
\\
\hline
{\textcolor{blue}{\bf T5 (\textit{With long prompt})}:} How does purchasing power parity help in economics? Read more about how does purchasing power parity help in economics?
\\
{\textcolor{blue}{\bf T5 (\textit{With short prompt})}:} What is the purpose of purchasing power?
\\
{\textcolor{blue}{\bf T5 (\textit{Without prompt})}:} What is the importance of a country's development index? How does it help the poor?
\\
\hline
{\textcolor{blue}{\bf MBART (\textit{With long prompt})}:} What does rupee purchasing power parity do for people living in economics?
\\
{\textcolor{blue}{\bf MBART (\textit{With short prompt})}:} What is the importance of purchasing power in daily living standards between two countries?
\\
{\textcolor{blue}{\bf MBART (\textit{Without prompt})}:} Why is it important to read a country-based economy pattern? Give example of poverty.
\\
\hline
{\textcolor{blue}{\bf BART (\textit{With long prompt})}:} How does purchasing power parity help improve economy parity? and why is it good?
\\
{\textcolor{blue}{\bf BART (\textit{With short prompt})}:} What is the importance of purchasing power in daily living? Purchasing power is important?
\\
{\textcolor{blue}{\bf BART (\textit{Without prompt})}:} What is the importance of ease of living standards in simple words, and why is it important?
\\
\hline
\hline
\end{tabular}
\end{table}

\begin{table}
\caption{ A data sample related to History in our \textit{EduProbe} dataset, along with corresponding questions generated by different LLMs using various prompt variations.}
\label{tab:t2}
  \footnotesize
  \begin{tabular}{|p{0.46\textwidth}|}
\hline
\hline
{\textcolor{red}{\bf Context}:} During the medieval period in India, Islamic rulers held significant power, leading to the blending of Indian and Islamic cultures, which can still be observed in the architecture and artwork created at that time. The country was governed and administered by notable rulers who made exceptional contributions in the fields of art, culture, literature, and architecture during this period.
\\ 
{\textcolor{Green}{\bf Long Prompt}:} the contribution of the medieval period to Indian history
\\ 
{\textcolor{Green}{\bf Short Prompt}:} medieval period
\\ 
{\textcolor{Green}{\bf Gold Standard Question}:} What is the contribution of the medieval period to Indian history?
\\ 
\hline
{\textcolor{blue}{\bf Davinci (\textit{\textit{\textit{with long prompt}}})}:} How did the convergence of Indian and Islamic cultures during the medieval period in India influence not only the realms of art, culture, literature, and architecture but also the broader socio-political fabric and historical trajectory of the nation?
\\
{\textcolor{blue}{\bf Davinci (\textit{\textit{\textit{with \textit{short prompt}}}})}:} How did the dynamic interplay between Indian and Islamic influences during the medieval period shape not only the artistic and architectural landscape but also the social, religious, and intellectual fabric of India, leaving a lasting impact on its history and identity?
\\
{\textcolor{blue}{\bf Davinci (\textit{\textit{\textit{without prompt}}})}:} How did the cross-cultural exchange between Indian and Islamic civilizations during the medieval period in India result in a unique synthesis of artistic, architectural, literary, and cultural expressions, and what enduring legacies did these contributions leave on the country's cultural heritage?
\\
\hline
{\textcolor{blue}{\bf ChatGPT (\textit{\textit{\textit{with long prompt}}})}:} What notable contributions were made to Indian history during the medieval period in terms of art, culture, literature, and architecture, reflecting the blending of Indian and Islamic influences?
\\
{\textcolor{blue}{\bf ChatGPT (\textit{\textit{\textit{with \textit{short prompt}}}})}:} What impact did the significant influence of Islamic rulers during the medieval period have on the architecture, artwork, governance, and cultural development in India?
\\
{\textcolor{blue}{\bf ChatGPT (\textit{\textit{\textit{without prompt}}})}:} What were the key cultural and artistic developments that emerged during the medieval period in India under the influence of Islamic rulers, and how did these contributions shape the architecture, artwork, literature, and overall cultural landscape of the time?
\\
\hline
{\textcolor{blue}{\bf Pegasus (\textit{\textit{\textit{with long prompt}}})}:} What is the contribution of the medieval period to Indian history?
\\
{\textcolor{blue}{\bf Pegasus (\textit{\textit{\textit{with \textit{short prompt}}}})}:} What was the impact of the medieval period on Indian culture, politics and economy? Give examples.
\\
{\textcolor{blue}{\bf Pegasus (\textit{\textit{\textit{without prompt}}})}:} What were the major events and developments that took place in the medieval period in India? Give examples.
\\
\hline
{\textcolor{blue}{\bf T5 (\textit{\textit{\textit{with long prompt}}})}:} What is the contribution of the medieval period to Indian history? Have you ever wondered what is the contribution of the medieval period to Indian history?
\\
{\textcolor{blue}{\bf T5 (\textit{\textit{\textit{with \textit{short prompt}}}})}:} What were the main events of the medieval period in India that took place during the Islamic period?
\\
{\textcolor{blue}{\bf T5 (\textit{\textit{\textit{without prompt}}})}:} What happened in the medieval period in India, which saw a strong control of Islamic rulers?
\\
\hline
{\textcolor{blue}{\bf MBART (\textit{\textit{\textit{with long prompt}}})}:} What is the contribution of the medieval period to Indian history? What is the contribution of the medieval period?
\\
{\textcolor{blue}{\bf MBART (\textit{\textit{\textit{with \textit{short prompt}}}})}:} What were the major events in the medieval period in India that took place during the ancient period?
\\
{\textcolor{blue}{\bf MBART (\textit{\textit{\textit{without prompt}}})}:} What were the major events that took place in India during the Second World War (WW2)?
\\
\hline
{\textcolor{blue}{\bf BART (\textit{\textit{\textit{with long prompt}}})}:} What is the contribution of the medieval period to Indian history? and what is its significance in Indian history?
\\
{\textcolor{blue}{\bf BART (\textit{\textit{\textit{with \textit{short prompt}}}})}:} What was the medieval period in India? and what was its importance for shaping public opinion? 
\\
{\textcolor{blue}{\bf BART (\textit{\textit{\textit{without prompt}}})}:} What was the main difference between the medieval period in India and the Chalcolithic period?\\
\hline
\hline
\end{tabular}
\end{table}
\section{Analysis}\label{sec9}
Both Davinci and ChatGPT generate questions that differ from the gold standard questions in terms of character, unigram, bigram, or longest common subsequence-based overlap \cite{r3,r1,r5,r2}. Therefore, in terms of automated metrics (see Table \ref{tab:t3}), these two LLMs cannot show good performance. However, the questions generated by both these general-purpose LLMs (\textit{e.g.}, Davinci and ChatGPT) are of good quality, as shown by the results of the human evaluation criteria in Table~\ref{tab:t4}. Davinci and ChatGPT show superior performance in terms of human criteria like \textit{grammaticality}, \textit{relevance}, \textit{appropriateness}, \textit{complexity}, and \textit{novelty}. However, Davinci and ChatGPT fall short of the human baseline in terms of \textit{grammaticality}, \textit{appropriateness}, and \textit{relevance} under the different prompt settings. But it has been observed that Davinci and ChatGPT can generate \textit{novel} questions that are not present in the gold standard, thus giving them high scores for the \textit{novelty} metric under the \textit{long prompt} setting. Overall, Davinci emerges as the best performer, followed by ChatGPT, based on most human evaluation criteria. Furthermore, it has also been observed that Davinci and ChatGPT can generate \textit{complex} questions in the \textit{\textit{without prompt}} setting. The amount of cognitive effort required to answer a question generated by Davinci and ChatGPT in the \textit{without prompt} setting is significantly higher than the human baseline. Fine-tuned domain-specific LLMs like T5 and BART show good performance in terms of automated metrics because these LLMs generate questions that are closer to the gold standard questions in terms of character, unigram, bigram, and longest common subsequence matches. However, these fine-tuned domain-specific LLMs fall short of Davinci and ChatGPT in terms of human evaluation criteria.

We observe different sets of results under different prompt settings and a broad and diverse range of questions generated from the same context, thereby showcasing the utility of prompt-based QG techniques. It suggests that prompts definitely help to vary the quality of the generated questions which is observed both in the case of pre-trained general-purpose LLMs (\textit{e.g.}, Davinci and ChatGPT) and fine-tuned domain-specific LLMs (\textit{e.g.}, Pegasus, T5, MBART, and BART). 

Table \ref{tab:t_2} and Table \ref{tab:t2} show two data samples from our \textit{\textit{EduProbe}} test set and the corresponding questions generated by different LLMs under various prompt settings.
\section{Conclusions and Future Works}\label{sec10}
We introduced \textit{\textit{EduProbe}}, a dataset to create deep and diverse questions that are more educationally oriented in the context of school-level subjects. We explored different types of prompt-based techniques (\textit{e.g.}, \textit{long prompt}, \textit{short prompt}, and \textit{\textit{without prompt}}) to provide QG models additional guidance on what information to emphasize more when generating questions. The experiments demonstrate that T5 surpasses other models in all automated metrics. Pre-trained general-purpose LLMs such as Davinci exhibit superior proficiency in generating questions that excel in terms of \textit{grammaticality}, \textit{appropriateness}, \textit{relevance}, \textit{novelty}, and \textit{complexity}. Furthermore, Davinci and ChatGPT surpass the human baseline in terms of generating \textit{complex} questions, though they fall short of the human baseline in terms of generating \textit{grammatical}, \textit{appropriate}, \textit{relevant}, and \textit{novel} questions.

We aim to explore even larger language models (\textit{e.g.}, GPT-4) for QG in the future on our proposed \textit{EduProbe} dataset. Currently, we are creating manual prompts through our annotators that can be replaced by automatic keyphrase or span detection models. Additionally, there is a need to develop better automated metrics for measuring the quality of generated questions, as current metrics cannot fully capture the quality of generated questions. We also plan to fine-tune general-purpose LLMs like Davinci in the future. Although LLMs have shown good performance in generating questions, they are still not able to reach human-level performance in most cases. Therefore, further research is required in this direction.

\bibliographystyle{ACM-Reference-Format}
\bibliography{sample-base}









\end{document}